\documentclass[pmlr,twocolumn,10pt]{jmlr} % W&CP article

\usepackage{booktabs}
\usepackage{siunitx}

\usepackage{microtype}
\usepackage{multirow}
\usepackage{multicol}
\usepackage{mathtools}
\usepackage{amsmath}

\theorembodyfont{\upshape}
\theoremheaderfont{\scshape}
\theorempostheader{:}
\theoremsep{\newline}

\jmlrvolume{LEAVE UNSET}
\jmlryear{2022}
\jmlrsubmitted{LEAVE UNSET}
\jmlrpublished{LEAVE UNSET}
\jmlrworkshop{Conference on Health, Inference, and Learning (CHIL) 2022} % W&CP title

\title{Enriching Unsupervised User Embedding via Medical Concepts}
\author{
\Name{Xiaolei Huang}%\equal{These authors contributed equally} 
\Email{xiaolei.huang@memphis.edu}\\
\addr University of Memphis, United States
\AND
\Name{Franck Dernoncourt}%\footnotemark[1]
\Email{dernonco@adobe.com}\\
\addr Adobe Research, United States
\AND
\Name{Mark Dredze} \Email{mdredze@cs.jhu.edu}\\
\addr Johns Hopkins University, United States
}

\begin{document}
\maketitle

\begin{abstract}
Clinical notes in Electronic Health Records (EHR) present rich documented information of patients to inference phenotype for disease diagnosis and study patient characteristics for cohort selection.
Unsupervised user embedding aims to encode patients into fixed-length vectors without human supervisions.
Medical concepts extracted from the clinical notes contain rich connections between patients and their clinical categories.
However, existing \textit{unsupervised} approaches of user embeddings from clinical notes do not explicitly incorporate medical concepts.
In this study, we propose a concept-aware unsupervised user embedding that jointly leverages text documents and medical concepts from two clinical corpora, MIMIC-III and Diabetes. 
We evaluate user embeddings on both extrinsic and intrinsic tasks, including phenotype classification, in-hospital mortality prediction, patient retrieval, and patient relatedness.
Experiments on the two clinical corpora show our approach exceeds unsupervised baselines, and incorporating medical concepts can significantly improve the baseline performance.
\end{abstract}

\paragraph*{Data and Code Availability}
In this study, we experiment with two clinical datasets that are publicly available, Medical Information Mart for Intensive Care (MIMIC)-III~\citep{johnson2016mimic} and Clinical Trial Cohort Selection (Diabetes)~\citep{stubbs2019cohort}.\footnote{We denote the cohort selection dataset as Diabetes because patients share the same syndrome, diabetes. Diabetes data link: \url{https://portal.dbmi.hms.harvard.edu/projects/n2c2-nlp/}, MIMIC-III data link: \url{https://physionet.org/content/mimiciii/1.4/}.}
The data description and processing details is in Section~\ref{sec:data}.
Our experimental codes, supplementary analyses, and data access instructions are at \url{https://github.com/xiaoleihuang/UserEmb_Explainable}.

\section{Introduction}

% Clinical notes are patient narratives generated by medical staffs to describe and summarize conditions and symptoms of patients. Clinical labels including International Classification of Diseases (ICD) are human annotations to indicate disease diagnosis and patient characteristics.

\textit{User embedding} aims to learn fix-length vectors for users and maps all user information into a unified vector space~\citep{pan2019social, li2020survey, si2020patient}. 
Patient representations inferred by user embeddings have shown broad applications in medical diagnosis~\citep{choi2016multilayer, choi2017gram}, hospital stays~\citep{wang2019inpatient2vec}, and patient readmission~\citep{darabi2020taper}.
The supervised learning approaches require a large amount of annotated datasets, yet obtaining human annotations for clinical notes can be labor-intensive, time-consuming, and expensive.
Additionally, International Classification of Diseases (ICD) codes or other human clinical annotations have annotating issues including noisy~\citep{harutyunyan2019multitask},  error-prone~\citep{birman2005accuracy}, and low inter-rater agreement~\citep{stubbs2019cohort}. 
Quality of data annotations can impact the robustness of supervised embedding models.

% unsupervised user embedding
\textit{Unsupervised} user embedding learns patient representations from clinical notes without human supervision.
The embedding aims to capture semantic relations between patients and their clinical records.
To represent clinical notes, models derive features from topic model~\citep{blei2003latent}, paragraph vectors~\citep{le2014distributed}, word2vec embeddings~\citep{mikolov2013distributed}.
Finding hidden structures to distinguish patients is a challenge for unsupervised user embeddings to compress patient features from clinical notes.
Unsupervised approaches~\citep{sushil2018patient, lei2018effective} utilize autoencoders by feature reconstruction, while ignoring other structures like medical concepts.

% Skip-gram~\citep{amir2017quantifying} and autoencoders~\citep{miotto2016deep, lei2018effective} are two unsupervised methods to compress features to fixed-length patient representations.

Medical \textit{concepts}, which consist of one or more tokens, are basic description units for medical information, such as disease symptom and clinical drug.
The concepts contain rich information of medical ontology that owns strong connections with phenotype inference.
Medical concept embeddings treat concepts as special tokens or phrases and encode them as word vectors. 
The representations are effective in downstream tasks of health diagnosis~\citep{choi2016multilayer, choi2020learning} and readmission forecast~\citep{choi2017gram, zhang2020learning}.
However, unsupervised user embeddings have not explicitly consider medical concepts, which can lose semantic information of clinical conditions.

In this study, we propose a \textit{concept-aware user embedding} (CAUE) model that jointly handles clinical notes and medical concepts under the multitask learning framework.
Our unsupervised method applies contrastive learning~\citep{logeswaran2018an, chen2020big} and negative sampling to leverages two levels of information sources, patient-document and patient-concept.
Patient-document task encodes sequential information of patients to guide models recognizing patients of clinical notes, and patient-concept task treats medical concepts as a medium point to constrain patients.
To model unusually lengthy clinical documents, we propose a data augmentation approach, random split, to cut each note into random sizes of snippets.
We evaluate on multiple clinical datasets with two extrinsic tasks, phenotype and mortality predictions, and two intrinsic tasks, patient relatedness and retrieval.
Research suggests that intrinsic evaluations of embedding models can reduce parameter biases of extrinsic evaluations for controlling fewer hyperparameters~\citep{schnabel2015evaluation}. 
% Intrinsic evaluations require no extra parameters of training external classifiers comparing to extrinsic evaluation approaches.
% Additionally, our intrinsic evaluations fit for patients who have a large number of clinical labels instead of limiting the label size~\citep{amir2017quantifying, purushotham2018benchmarking, harutyunyan2019multitask}.
The four evaluations show that our proposed approach achieves the best performance overall and highlight the effectiveness of medical concepts.
Our ablation analysis shows that integrating medical concepts can generally improve effectiveness of baseline models.

\section{Data}
\label{sec:data}

\begin{table*}[t!]
\centering
% \resizebox{\textwidth}{!}{
    \begin{tabular}{c||ccc||cccc||cc}
    \multirow{2}{*}{Dataset} & \multicolumn{3}{c||}{Document Statistics} & \multicolumn{4}{c}{User Stats} & \multicolumn{2}{c}{Concept Stats}\\
     & Doc & Vocab & Token-stats & User & Age & F & U-label & Concept & Type\\\hline\hline
    Diabetes & 1265 & 34592 & 2426, 483, 42 & 288 & 63.13 & 0.45 & 10 & 68938 & 89 \\
    MIMIC-III & 54888 & 390237 & 7522, 1263, 50 & 48807 & 62.47 & 0.44 & 276 & 10761211& 94
    \end{tabular}
% }
\caption{Data summary of Diabetes and MIMIC-III by documents, patients, and concepts. The ``Token-stats'' includes maximum, median and minimum lengths of processed clinical notes. We present values of patient count (User), averaged age (Age), gender ratio of female (F) and male (M), number of unique labels (U-label), number of concepts (Concept), and number of unique concept types (Type).}
\label{tab:data}
\end{table*}

% Introduce the two datasets, compare the datasets with other clinical datasets, compare the clinical dataset with other types of datasets, like the social media documents.
In this section, we introduce the two clinical data (Diabetes and MIMIC-III) and report data processing steps.
We obtain the Diabetes data~\citep{stubbs2019cohort} from Track 1 of the 2018 National NLP Clinical Challenges (n2c2) shared task, which contains a collection of longitudinal patient records, with 2 to 5 text documents per patient.
The clinical documents summarize diagnosis results and describe clinical conditions indicating if patients have coronary artery disease.
Each patient in the Diabetes data has 13 selection criteria annotations, which follow ClinicalTrials.gov and relate to diabetes and heart disease. 
MIMIC-III is a relational database that collects a large set of intensive care unit (ICU) patients from the Beth Israel Deaconess Medical Center between 2001 and 2012.\footnote{We access the data at \url{https://physionet.org/content/mimiciii/1.4/}. MIMIC has different versions, and we experiment with the stable version, MIMIC-III 1.4. Its release year is 2016.}
The data includes various patient information such as demographics, laboratory results, mortality, and clinical notes. 
Clinical notes are unstructured narrative documents describing patients' ICU stays, including radiology and discharge summaries. 
Like Diabetes, each MIMIC-III document has a list of medical annotations, International Classification of Disease, 9th Edition (ICD-9).
The ICD-9 annotations indicate symptoms and disease types of patients during their ICU stays.
Both data corpora have been de-identified following Health Insurance Portability and Accountability Act (HIPAA) standards, and any dates were time-shifted by a random amount.

% extract medical concepts
We use MetaMap~\citep{aronson2010overview} to extract medical concepts by the Unified Medical Language System Metathesaurus (UMLS)~\citep{bodenreider2004unified}.\footnote{While we have alternative toolkits, such as cTAKES \citep{savova2010mayo} and CLAMP toolkit~\citep{soysal2017clamp}, however, they did not adapt to the new authentication system of UMLS when this study started.}
MetaMap extracts medical concepts with their Concepts Unique Identifier (CUI) linked to the UMLS, confidence scores, and corresponding medical types.
To improve accuracy and reduce the ambiguity of extracted concepts, we enable the word sense disambiguation (WSD) module. 
The WSD module checks if concepts are semantically consistent with surrounding texts and automatically removes ambiguous entities.
Each concept may have one or more tokens and associate with at least one concept type.
We lowercase all extracted medical concepts and empirically removed non-symptom-related entities, including digits-only sequences, language, and temporal concepts.\footnote{MetaMap concept types: \url{https://metamap.nlm.nih.gov/Docs/SemanticTypes_2018AB.txt}}

In this study, we preprocess medical annotations and clinical notes for both corpora.
For MIMIC-III, we follow previous work~\citep{mullenbach2018explainable} to keep discharge summary, which merges enough details of a single admission for each patient.
ICD-9 labels for the MIMIC-III have over 15K different codes, which are highly dimensional and sparse.
To reduce sparsity and noise, we follow previous work~\citep{harutyunyan2019multitask} to merge ICD-9 codes according to the Health Cost and Utilization (HCUP) Clinical Classification
Sofware (CCS).\footnote{\url{https://www.hcup-us.ahrq.gov/toolssoftware/ccs/ccs.jsp}} 
The predefined HCUP-CCS schema clusters ICD-9 procedure, billing, and diagnostic codes into mutually exclusive, largely homogeneous disease categories.
We follow previous work~\citep{harutyunyan2019multitask} to drop any patients who are younger than 18 years old.
Two domain experts annotated each patient from the Diabetes data by 13 different clinical criteria.
We empirically drop three of them because their inter-rater agreements are fewer than 0.5.\footnote{The agreement rate of ENGLISH, MAKES-DECISIONS, and KETO-1YR are 0.46, 0.31, -0.1, respectively. Note that the main reason that causes a negative agreement score of KETO-1YR is the skewed distributions of annotations.}
We apply the same preprocessing steps for clinical notes.
Each patient associates with one or more notes and their medical annotations.
The temporal shift is an attribute of the in-hospital records, MIMIC-III. 
To reduce temporal impacts and noises, we isolate a patient's records per visit as a separate patient.
For each note, we lowercase all tokens, tokenize sequences by NLTK~\citep{bird2004nltk}, remove person and hospital names, replace numbers and dates with placeholders ([NUM] and [DATE]), and remove repeated punctuation.
We empirically drop any notes with less than 40 tokens.

% Brief discussion about data statistics
We summarize data statistics in Table~\ref{tab:data}.
MIMIC-III shows more varied statistics of clinical notes than the Diabetes dataset. 
For example, the range of token counts for MIMIC-III data is broader than Diabetes (50 to 7522 vs. 42 to 2426), and MIMIC-III has a larger vocabulary.
Both datasets have approximately similar demographic distributions of age and gender.
MIMIC-III data has more complex clinical annotations, while Diabetes data has a maximum of 10 labels for each patient.
Comparing to health data from social media, such as Twitter, the clinical corpora have more senior user groups and more variations in documents, patients, and annotation information.
For example, while the Twitter suicide users~\citep{amir2017quantifying} only has binary labels, MIMIC-III data has 276 unique phenotypes.

% \begin{figure*}[ht]
% \centering
% \includegraphics[width=0.497\textwidth]{images/diabetes_quant.pdf}~~~
% \includegraphics[width=0.497\textwidth]{images/mimic-iii_quant.pdf}
% \caption{Phenotype classification performance measured by F1 score. We conduct experiments on Diabetes (left) and MIMIC-III (right). We show three feature sets, n-gram (blue bar), concept (orange bar) and a combination of n-gram and concept (green bar).}
% \label{fig:quant}
% \end{figure*}

\subsection{Privacy and Ethical Considerations}

We do not release any clinical data associate with patient identities due to privacy and ethical considerations.
Instead, we have released our code and provided detailed instructions to replicate our analysis and experiments.
To protect user privacy, we have followed corresponding data agreements to ensure the proper data usage and experimented with de-identified data.
Our experiments do not store any patient data and only use available text documents for research demonstrations.
Except, our experiments use anonymized patient ID and clinical notes for training and evaluating user embeddings.
% This work has been approved by Institutional Review Board (IRB).

\section{Concept-Aware User Embedding (\textit{CAUE})}
\label{sec:model}

\begin{figure*}[htp]
\centering
\includegraphics[width=0.743\textwidth]{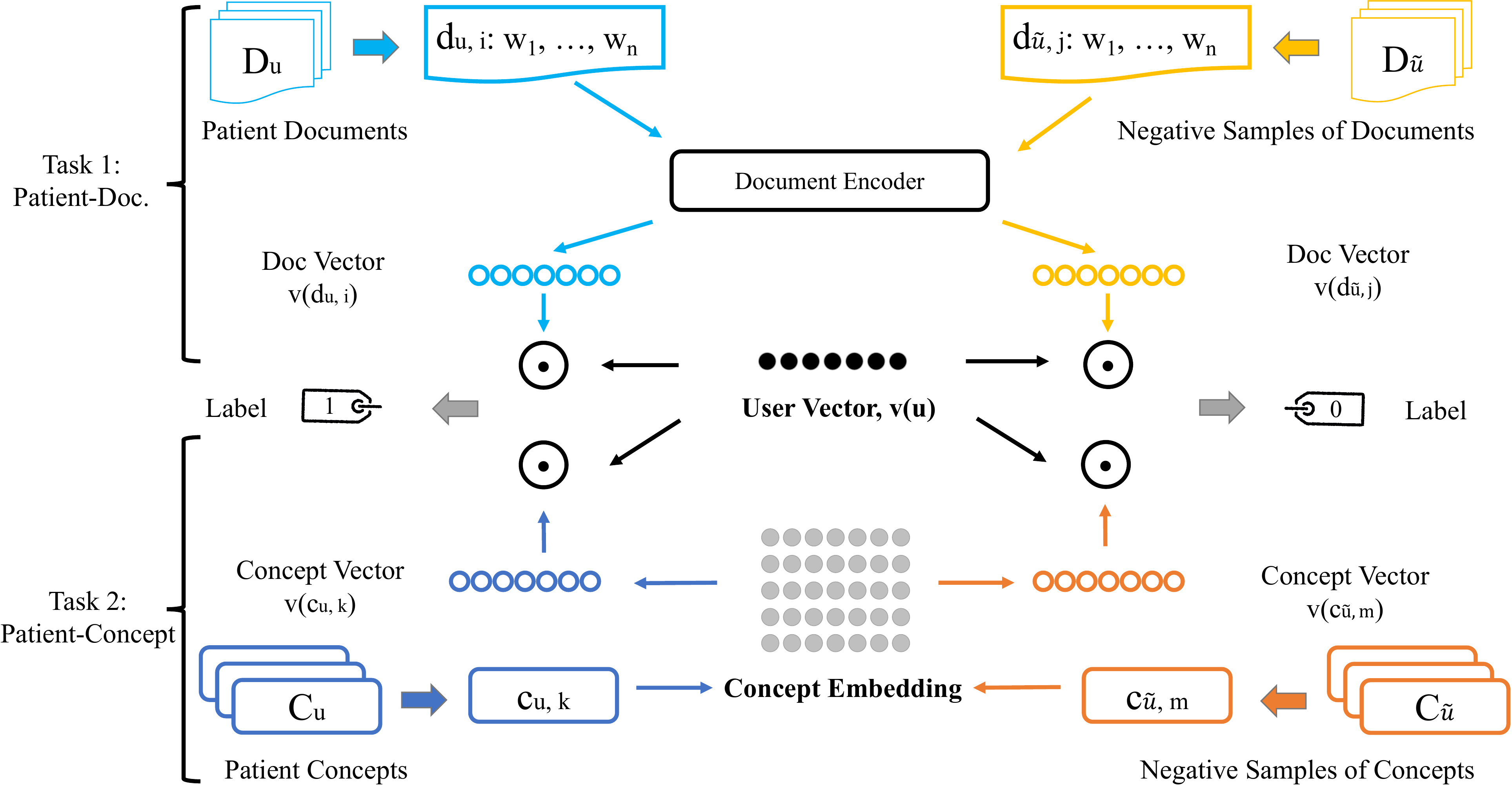}
\caption{CAUE illustrations. All tasks center with the the patient, $u$. The arrows and their colors refer to the input directions and sources respectively. The $\bigodot$ is a similarity measurement.}
\label{fig:diagram}
\end{figure*}

In this section, we propose an \textbf{unsupervised} concept-aware user embedding (\textit{CAUE}) that jointly models clinical notes and medical concepts under a multitask learning framework.
Methods~\citep{si2020deep} to train unsupervised user embedding from clinical notes primarily focus on bag-of-words features without explicitly capturing semantic information of medical concepts.
\citep{steinberg2021language} utilizes medical codes of clinical records as input features to derive patient representations sequentially, while our focus is to model sequential language features, clinical notes.
\citep{dligach2018learning} obtains user embeddings by training a supervised classifier using concept embeddings. However, our approach jointly models document and concept embeddings in an unsupervised manner.
% A close work to ours only trained user embeddings by predicting if social media users co-occurred with sampled words~\citep{amir2017quantifying}.
We propose two joint tasks: modeling language sequential dependency by neural document encoders (task 1: patient-document) and leveraging connections between medical concepts and patients (task 2: patient-concept).
We present the model architecture and learning process in Figure~\ref{fig:diagram}.

\subsection{Task 1: Patient-Document}

The first task is to enforce models to understand contextual and sequential information of clinical notes so that models can summarize patients.
Instead of training individual tokens, this task aims to model connections between users with documents.
Given $U$ be the list of patients, $D_u$ be a list of clinical notes that describe the patient $u$, and we have $u \in U$, we can formalize our goal by maximizing the conditional probability of $P(D_u|u)$.
Let $d_{u, i}$ as a clinical document (index as $i$) that describes the patient $u$ clinical information, and each clinical note is independent with other notes in the collection $D_u$, our user-document task aims to maximize the following probability:

\begin{equation}
\displaystyle P(D_u|u) = \sum_{i=1}^{|D_u|} log P(d_{u, i}|u)
\end{equation}
To maximize the probability, we convert the task to a prediction task. Given a user, a model predicts a probabilistic distribution of a user vector over the patient's clinical notes so that the clinical note summarizes its target patient.

% \paragraph{Contrastive Learning}
Considering the large size of the patients and clinical notes, we approximate our optimization objectives by the negative sampling and contrastive learning.
The negative sampling is to randomly choose different documents from the existing corpus besides the target clinical document.
The \textit{contrastive learning}~\citep{logeswaran2018an, chen2020big} is to predict if a sampled document or sentence is a context of the target.
Previous study~\citep{logeswaran2018an} predicts the context sentence from several sampled or generated counterfactuals, which are stylistically similar to the context sentence but express different meanings.
In this study, we treat patients as our target and clinical documents as context so that learning process will align patient representations toward their correct context, clinical documents.
To generate counterfactuals, we choose two types of information: document and token.
For the document level, we sample 3 snippets as counterfactuals by randomly choosing documents from the other patients.
For the token level, we randomly replace tokens in the original document, where the sampling tokens are the vocabulary of each corpus. 
Instead of predicting each clinical document over all patients, the negative sampling enables model predictions over a few samples.
This converts our prediction task into predicting if our model can identify if a text document describes the target patient or not.
Let $V$ as embeddings for documents, users and concepts, we can derive a document vector $v(d)$ for the document $d_u$, a user vector $v(u)$ for the patient $u$, and a concept vector $v(c)$ for the concept $c$.
Then we can treat the task as a binary classification problem and minimize loss values by the binary cross-entropy: 

\begin{equation}
\begin{split}
\mathcal{L}(u, d) & = -log(\sigma(v(u) \cdot v(d_u))) \\
& - log(1-\sigma(v(u) \cdot v(d_{\Tilde{u}})))
\end{split}
\end{equation}
where $\sigma$ is a non-linear function and $d_{\Tilde{u}}$ refers to noise documents.
We use the sigmoid function as our non-linear function to normalize values of the dot production.
The binary classification task will predict if a sampled clinical document describes correctly the target user, where one means correct match and 0 means incorrect descriptions.

We utilize a document encoder to derive document vectors $v(d)$ to learn language dependency across sequential tokens. 
Let a document $d$ has a sequence of words, $d = \{w_1, w_2, ..., w_n\}$.
Then we can derive document vectors by $v(d) = f_d(v(w_1), ..., v(w_n))$, where $f_d$ is the document encoder and $v(w)$ is a word vector.
Finally, our document encoder can learn sequential information of each clinical note into a unified vector, $v(d)$.

\subsection{Long Text Modeling via Random Split}
The clinical notes have a much longer document length (e.g, 7522 in our case) comparing to Twitter messages.
Such a long sequence prevents effectively training neural models and fitting GPU memory because of sequence length limits.
For example, the BERT~\citep{devlin2019bert} has a limitation of 512 tokens.
We solve the problem by splitting each document longer than a length limit into multiple snippets with a random size, which is between 200 and 512 tokens.
This simulates clinical settings in real-world that physicians can recognize their patients with partial symptom descriptions.
% Padding operations ensure training snippets have the same document length.
% Empirically, we find that longer snippets have lower loss values.
% For example, if a document has 900 tokens, to obtain two snippets with an equal size: we first split the first 512 tokens as one snippet; next, we count 512 tokens from the end of the document to get the second snippet; finally, we can have two snippets with equal size.

\subsection{Task 2: Patient-Concept}

The second task is to \textit{enrich} user representations by extracted concepts from clinical notes. 
On the one hand, concepts are identities of users if the concepts co-occur with the users.
On the other hand, concept entities can connect users who share similar medical symptoms and separate users who do not share similar information.
Given that $C_u$ be a list of extracted medical concepts from the patient's clinical notes and we have $c \in C$, following task 1, we can formalize our goal as maximizing the conditional probability of $P(C_u|u)$.
Let $c_{u, k}$ as a medical concept (index as $k$) that occurs in the clinical notes of the patient $u$, our patient-document task aims to maximize the following probability:

\begin{equation}
\displaystyle P(C_u|u) = \sum_{k=1}^{|C_u|} log P(c_{u, k}|u)
\end{equation}

To reduce computational complexity, we employ negative sampling to approximate our task goal. 
The negative sampling can help us convert the task to a binary prediction task that checks if a sampled concept exists in the patient's clinical notes.
If a sampled concept exists in the target patient's clinical notes, the label is 1; otherwise, the prediction label is 0.
The classification problem minimizes loss values by the binary cross-entropy:

\begin{equation}
\begin{split}
\mathcal{L}(u, c) & = -log(\sigma(v(u) \cdot v(c_u))) \\
& - log(1-\sigma(v(u) \cdot v(c_{\Tilde{u}})))
\end{split}
\end{equation}
, where $c_{\Tilde{u}}$ refers to negative samples of noise documents.
Following the task 1, we use the sigmoid function as our non-linear function to normalize values of the dot production.
For each concept, we sample another 3 noise concept sets which do not co-occur with the target patient.

% \subsection{Masked Language Modeling}

\subsection{Joint Learning}
We optimize user embeddings by patient-document, patient-concept, and masked language modeling (mlm) tasks in parallel.
The joint tasks can incorporate document and concept information into user representations. 
For the mlm, we follow pre-training tasks of the BERT~\citep{devlin2019bert} to randomly masked out 15\% tokens \textit{only if} the CAUE models are using the BERT as neural feature extractors.
We sum the total loss value by:

\begin{equation}
\begin{split}
\mathcal{L} & = \lambda \mathcal{L}(u, c) + (1-\lambda)\mathcal{L}(u, d) + \alpha\mathcal{L}_{mlm}
\end{split}
\end{equation}
where $\lambda$ and $\alpha$ are weight factors to balance different tasks. 
We empirically set $\lambda$ as 0.3 and $\alpha$ as 0.03.

\section{Experiments}
\label{sec:exp}

In this section, we present experimental settings of our approach and baselines.
To keep consistency, we fix embedding dimensions as 300 for user and concept embeddings.
We include more implementation details in our appendix.

\begin{table*}[ht!]
\centering
\resizebox{1\textwidth}{!}{
    \begin{tabular}{c||cc||c||cc||cc}
    \multirow{2}{*}{Models} & \multicolumn{2}{c||}{Phenotype Prediction} & Mortality & \multicolumn{2}{c||}{Patient Relatedness} & \multicolumn{2}{c}{Retrieval} \\
              & Diabetes & MIMIC-III & MIMIC-III & Diabetes & MIMIC-III & Diabetes & MIMIC-III\\\hline\hline
    \multicolumn{8}{c}{Baselines}\\\hline\hline
    word2user & 0.574 & 0.321 & 0.861 & 0.530 & 0.809 & 0.374 & 0.065\\
    doc2user & 0.288 & 0.235 & 0.425 & 0.573 & 0.253 & 0.344 & 0.132\\
    dp2user & 0.416 & 0.315 & \textbf{0.868} & 0.542 & 0.829 & 0.318 & 0.113\\
    suisil2user & 0.413  & 0.274 & 0.424 & 0.542 & 0.830 & 0.313 & 0.132\\
    usr2vec & 0.556 & 0.158 & 0.423 & 0.355 & 0.252 & 0.349 & 0.137\\
    siamese2user & 0.590 & 0.253 & 0.750 & 0.410 & 0.218 & 0.360 & 0.116\\\hline\hline
    \multicolumn{8}{c}{Ours} \\\hline\hline
    CAUE-GRU & \textbf{0.635} & \textbf{0.369} & 0.817 & \textbf{0.292} & \textbf{0.205} & 0.386 & 0.139 \\
    CAUE-BERT & 0.593 & 0.339 & 0.792 & 0.337 & 0.235 & \textbf{0.396} & \textbf{0.142}
    \end{tabular}
}
\caption{Two types of performance evaluations: supervised (phenotype and mortality classification) and unsupervised (patient relatedness and retrieval). For the patient relatedness, lower scores are better.}
\label{tab:performance}
\end{table*}

\subsection{CAUE Settings}

We treat neural models as feature extractors and experiment with two types of neural document encoders, Gated Recurrent Unit (GRU)~\citep{cho2014properties} and BERT~\citep{devlin2019bert}.
We denote the two types of encoders as \textit{CAUE-GRU} and \textit{CAUE-BERT} in the following.
Each type of encoder reads through sequential tokens and generates document representations.
We set maximum epochs as 15, batch size as 16, and vocabulary size as 15,000 for both token and concept.

\textbf{CAUE-GRU} utilizes one GRU layer to encode each document.
To incorporate more contextual information, we employ a bi-directional GRU layer.
Its output is a concatenation of bi-directional hidden vectors.
We set the dimension of the GRU layer as 300 and apply a dropout with 0.2 on outputs of the GRU layer.
We use a pre-trained word embedding~\citep{zhang2019biowordvec} to encode tokens into vectors.
The embedding was pre-trained on PubMed and MIMIC-III corpora.
While extensive research methods have explored how to train concept embeddings~\citep{choi2016multilayer, zhang2020learning}, our study uses the concepts as mediums to learn hidden patterns to better separate patients. 
We induce an initial concept embedding for (multi-token) concepts by averaging constituent word vectors.
% We freeze word embeddings during training steps, but we set the concept embedding as trainable.
We optimize the model using RMSprop~\citep{hinton_srivastava_swersky} with a learning rate $1e-4$.

\textbf{CAUE-BERT} encodes document into fixed-length vectors via BERT~\citep{devlin2019bert}.
The BERT model encodes each document by a hierarchical transformer~\citep{vaswani2017attention} layers.
We add a fully connected layer on top of the encoded \texttt{[CLS]} token from the last layer of the BERT model to map 768-dimensional vectors into 300 dimensions.
To get document vectors, we pad input documents into a fixed-length, which is a hyperparameter.
As initializing the concept embedding with other pre-trained embeddings does not align with the BERT in the vector space,
we treat concepts as short sentences and utilize the BERT model to obtain concept embeddings.
We experimented with multiple pre-trained BERT models~\citep{peng2019transfer, alsentzer2019publicly, lee2019biobert}. However, we did not find significant improvements with the different choices. 
We adopt AdaW~\citep{loshchilov2018decoupled} to optimize model parameters with a learning rate of $3e-5$.
We pad documents with less than 512 tokens to ensure every input document has an equal size.

\subsection{Baselines}
In this study, we compare our approach with six unsupervised user embeddings.
To keep fair comparison, we keep the same embedding dimensions, run experiments with three times, and average model performances.
We use the same pre-trained embeddings~\citep{zhang2019biowordvec, alsentzer2019publicly} throughout the baselines.
% In this study, we compare our approach with six unsupervised user embeddings: word2user~\citep{benton2016learning}, doc2user~\citep{ding2017multi}, suisil2user~\citep{sushil2018patient}, dp2user~\citep{miotto2016deep}, usr2vec~\citep{amir2017quantifying}, and siamese2user~\citep{mueller2016siamese, huang2021user}.

\textbf{word2user} represents users by averaging over aggregated word
representations~\citep{benton2016learning}.
We calculate a user representation by averaging embeddings of all tokens in the clinical notes of each patient. 
We use the pre-trained word embedding~\citep{zhang2019biowordvec} to encode tokens.

\textbf{doc2user} applies paragraph2vec~\citep{le2014distributed} to obtain user vectors. 
We implemented the User-D-DBOW model, which achieved the best performance in the previous work~\citep{ding2018predicting}. 
The model trains the paragraph2vec for 10 epochs and left other parameters with default values in the Gensim~\citep{rehurek2010software}.
We append clinical documents of each patient as a single document. Then the User-D-DBOW model can derive a single user vector from the aggregated document.

\textbf{suisil2user} combines the doc2user with medical concepts to obtain patient vectors~\citep{sushil2018patient}. 
We reuse doc2vec to derive document features and follow the previous work to extract concept features. 
We first derive TF-IDF weighted scores of medical concepts for each patient and train an autoencoder to reduce the dimensions of concept features.
The autoencoder learns user embeddings by compressing features inputs into intermediate embeddings and reconstructing them back feature inputs.
Following the previous work, we set the sigmoid as the default activation function.
We add a 0.05 noise rate and train the autoencoder for ten epochs.
The document vectors concatenate with dimension-reduced concept vectors to generate patient representations.

\textbf{dp2user} generates patient representations by applying an autoencoder on extracted topic features~\citep{miotto2016deep}. 
We train Latent Dirichlet Allocation (LDA)~\citep{blei2003latent} on the patient level and apply the LDA on clinical notes to derive topic features.
We set the number of topics as 300, train the model for 10 epochs, and leave the rest of the parameters as their default values in Gensim~\citep{rehurek2010software}.
We train a four-layer autoencoder with a noise score of 0.05 to reduce dimensions of topic features.

\textbf{usr2vec} trains user embeddings by predicting if users authored sampled tokens~\citep{amir2017quantifying}.
The predictive goal is to measure if sampled words co-occur with a patient in clinical notes in terms of clinical notes.
To make a fair comparison, instead of randomly initializing word embeddings from scratch, we initialized embedding weights as the same pre-trained embeddings~\citep{zhang2019biowordvec} in our approach.
% follow the existing work~\citep{amir2017quantifying} by training
We train the model for 10 epochs, set the number of negative samples as three.

\textbf{siamese2user} utilizes a Siamese~\citep{mueller2016siamese} recurrent architecture~\citep{huang2021user} that learns patient representations on the document level. The Siamese model measures similarities between two sentences. We encode each clinical note via a bidirectional GRU, which is beyond the token-level usr2vec. For the patient, we apply a feed-forward network on the patient vectors and optimize the model by distances between document and patient vectors.

\subsection{Extrinsic 1: Phenotype prediction}

% \begin{table*}[ht]
% \centering
% \begin{tabular}{c||ccccc||ccc}
% \multirow{2}{*}{Dataset} & \multicolumn{5}{c||}{Baselines} & \multicolumn{3}{c}{Our Approach} \\
%  & word2user & doc2user & suisil2user & dp2user & usr2vec & siamese2user & CAUE-GRU & CAUE-BERT \\\hline\hline
% Diabetes & 0.574 & 0.288 & 0.413 & 0.416 & 0.556 & 0.590 & \textbf{0.635} & 0.573 \\
% MIMIC-III & 0.321 & 0.235 & 0.274 & 0.315 & 0.158 & 0.253 & \textbf{0.369} & 0.339
% \end{tabular}
% \caption{Performance evaluations of phenotype classification task by mean average precision (MAP). We highlight the best performance for each dataset. A higher score means a better performance.}
% \label{tab:phenotype}
% \end{table*}

% \begin{table}[ht]
% \centering
% \begin{tabular}{c|cc}
%     Models & Diabetes & MIMIC-III \\\hline
%     \multicolumn{3}{c}{Baselines} \\\hline
%     word2user & 0.574 & 0.321 \\
%     doc2user & 0.288 & 0.235 \\
%     suisil2user & 0.413 & 0.274 \\
%     dp2user & 0.416 & 0.315 \\
%     usr2vec & 0.556 & 0.158 \\\hline
%     \multicolumn{3}{c}{Our Approach} \\\hline
%     siamese2user & 0.590 & 0.253 \\
%     CAUE-GRU & \textbf{0.635} & \textbf{0.369} \\
%     CAUE-BERT & 0.573 & 0.339
% \end{tabular}
% \caption{Performance evaluations of phenotype classification task by mean average precision (MAP). We highlight the best performance for each dataset. A higher score means a better performance.}
% \label{tab:phenotype}
% \end{table}

We train classifiers with user embeddings as input features to infer phenotype (ICD-9 codes).
Each patient has an encoded set, a one-hot vector, which is a length of the unique labels.
The task is to evaluate how accurately a classifier can predict the clinical labels of each patient.
We use the embedding to encode patients and feed patient vectors to the logistic regression (LR) classifier for each user embedding.
We optimize the LR classifier by Adam~\citep{kingma2014adam} and L2-regularization.
We set the learning rate as 0.001 and the regularization score as 0.01.
We use 5-fold cross-validation to evaluate classification performance. 
Each fold splits the Diabetes and MIMIC-III data based on their patient counts by 80\% as a training set and 20\% as a test set.
Finally, we report mean average precision (MAP) for model evaluation.

We present our classification results in Table~\ref{tab:performance}.
Comparing to the baselines, LR classifiers trained by our proposed approach (CAUE-GRU) achieve the best performance across the two datasets. The results highlight that our method can capture patient variations and augment phenotype inference via the joint optimization tasks. 
The only difference between the siamese2user and usr2vec methods is whether it is on document level.
We can find that siamese2user outperforms usr2vec, which indicates that learning unsupervised user embedding can benefit from sequential dependency beyond individual tokens.
% We can find that usr2vec performs poorly

\subsection{Extrinsic 2: In-hospital mortality prediction}

% \begin{table*}[ht]
% \centering
% \begin{tabular}{c||ccccc||ccc}
% \multirow{2}{*}{Dataset} & \multicolumn{5}{c||}{Baselines} & \multicolumn{3}{c}{Our Approach} \\
%  & word2user & doc2user & suisil2user & dp2user & usr2vec & siamese2user & CAUE-GRU & CAUE-BERT \\\hline\hline
% MIMIC-III & 0.861 & 0.425 & 0.424 & \textbf{0.868} & 0.423 & 0.750 & 0.817 & 0.762 
% \end{tabular}
% \caption{Performance evaluations of in-hospital mortality prediction by F1-macro. We highlight the best performance for each dataset. A higher score means a better performance.}
% \label{tab:mortality}
% \end{table*}

% \begin{table}[ht]
% \centering
% \begin{tabular}{c|cc}
%     Models & MIMIC-III \\\hline
%     \multicolumn{2}{c}{Baselines} \\\hline
%     word2user & 0.861 \\
%     doc2user & 0.425 \\
%     suisil2user & 0.424 \\
%     dp2user & \textbf{0.868} \\
%     usr2vec & 0.423 \\\hline
%     \multicolumn{2}{c}{Our Approach} \\\hline
%     siamese2user & 0.750 \\
%     CAUE-GRU & 0.817 \\
%     CAUE-BERT & 0.762
% \end{tabular}
% \caption{Performance evaluations of in-hospital mortality prediction by F1-macro. We highlight the best performance for each dataset. A higher score means a better performance.}
% \label{tab:mortality}
% \end{table}

In-hospital mortality is a binary classification task to predict if patients died during in-hospital stay based on the first 48 hours of an ICU stay. 
We follow the benchmark~\citep{harutyunyan2019multitask} to generate inpatient mortality labels.
The mortality annotations are only available for the MIMIC-III dataset.
We use 5-folds cross-validation to train LR classifiers and F1-score (macro) to evaluate the performance of mortality prediction as for the macro is less sensitive to imbalanced mortality labels.

We present mortality prediction results in Table~\ref{tab:performance}.
Our approach generally performs well among baselines. 
We can find that our close work usr2vec performs worse than siamese2user.
Training mechanisms of usr2vec is the poor performance on both classification tasks.
% The result demonstrates that modeling language sequential information can improve classification performance on the two datasets.
The usr2vec predicts relations between words and users on Twitter messages, which have much shorter documents.
The negative sampling space will be too sparse to find strong patterns that separate patients.
This highlights the unique challenge to derive unsupervised user embeddings on clinical settings.

\subsection{Intrinsic 3: patient relatedness}

% \begin{table*}[ht]
% \centering
% \begin{tabular}{c||ccccc||ccc}
% \multirow{2}{*}{Dataset} & \multicolumn{5}{c||}{Baselines} & \multicolumn{3}{c}{Our Approach} \\
%  & word2user & doc2user & suisil2user & dp2user & usr2vec & siamese2user & CAUE-GRU & CAUE-BERT \\\hline\hline
% Diabetes & 0.530 & 0.573 & 0.542 & 0.542 & 0.355 & 0.410 & \textbf{0.292} & 0.573  \\
% MIMIC-III & 0.809 & 0.253 & 0.830 & 0.829 & 0.252 & 0.218 & \textbf{0.205} & 0.253
% \end{tabular}
% \caption{Evaluation results of patient relatedness task by mean square error (MSE). We highlight the best performance for each dataset. Lower scores are better.}
% \label{tab:relatedness}
% \end{table*}

% \begin{table}[ht]
% \centering
% \begin{tabular}{c|cc}
%     Models & Diabetes & MIMIC-III \\\hline
%     \multicolumn{3}{c}{Baselines} \\\hline
%     word2user & 0.530 & 0.809 \\
%     doc2user & 0.573 & 0.253 \\
%     suisil2user & 0.542 & 0.830 \\
%     dp2user & 0.542 & 0.829 \\
%     usr2vec & 0.355 & 0.252 \\\hline
%     \multicolumn{3}{c}{Our Approach} \\\hline
%     siamese2user & 0.410 & 0.218 \\
%     CAUE-GRU & \textbf{0.292} & \textbf{0.205} \\
%     CAUE-BERT & 0.337 & 0.235
% \end{tabular}
% \caption{Evaluation results of patient relatedness task by mean square error (MSE). We highlight the best performance for each dataset. Lower scores are better.}
% \label{tab:relatedness}
% \end{table}

The patient relatedness task is to measure semantic relation or similarity between patients. 
Data to evaluate relatedness originates from lexical semantics.
SimLex-999~\citep{hill2015simlex} provides a list of word pairs with similarity scores from human annotators.
SimLex-999 checks if similarities of word vectors are close to human perceptions.
Inspired by this, we utilize an intrinsic evaluation that measures the quality of user embeddings by patient relatedness.
However, unlike word embedding evaluation, there is no human judgement for patient-level similarities.
Therefore, we adapt medical codes to measure patient similarities.
The patient relatedness assumes that the similarity between every two patients should be proportional to the similarity of their clinical labels:

\begin{equation}
    Similarity(u_1, u_2) \propto Similarity(l_1, l_2)
\end{equation}
, where $u1 \in U$, $u2 \in U$, $l_1$ and $l_2$ are patients' corresponding phenotype labels.
We use user embeddings to encode patients into vectors and encode the labels into one-hot vectors.
The dimension of each label vector is the number of unique clinical labels. For example, $l_1$ in MIMIC-III data has 276 dimensions, corresponding to number of unique phenotype labels in Table~\ref{tab:data}.
We use cosine similarity to measure patient relatedness between vectors.
\textit{Mean Squared Error} (MSE) is to measure differences between a pair of patients and a pair of label vectors:
\begin{equation}
MSE = \frac{1}{|U|}\sum_{i=1}^{|U|}(Sim(l_1, l_2) - Sim(u_1, u_2))^2
\end{equation}

We present our evaluation results of patient relatedness in Table~\ref{tab:performance}.
Overall our concept-aware user embedding (CAUE-GRU) achieves the best performance over baselines.
% The observations suggest that modeling sequential information can identify user variations and separate users according to their clinical conditions.
While the suisil2user achieves better performance than the doc2user on Diabetes, it fails on the MIMIC-III.
Furthermore, our approach yields good performance on the two datasets.
The result can highlight that how to encode concept features into user embeddings are critical for improving model robustness.

\begin{table*}[ht!]
\centering
\resizebox{\textwidth}{!}{
    \begin{tabular}{c||cc||c||cc||cc}
    \multirow{2}{*}{ + Concept} & \multicolumn{2}{c||}{Phenotype Prediction} & Mortality & \multicolumn{2}{c||}{Patient Relatedness} & \multicolumn{2}{c}{Retrieval} \\
     & Diabetes & MIMIC-III & MIMIC-III & Diabetes & MIMIC-III & Diabetes & MIMIC-III \\\hline\hline
    word2user & +7.6\% (.044) & +40.4\% (.130) & +4.8\% (.041) & +3.4\% (-.018) & +1.2\% (-.010) & -0.5\% (-.002) & +24.6\% (.016) \\
    doc2user & -43.4\% (-.125) & -32.3\% (-.076) & +2.8\% (.012) & +1.1\% (-.006) & +0.8\% (-.02) & +6.1\% (.021) & -3.0\% (-.004) \\
    dp2user & +22.6\% (.094) & +52.1\% (.164) & +3.8\% (.033) & +2.0\% (-.011) & +24.9\% (-.207) & +17.6\% (.056) & +18.6\% (.021) \\
    usr2vec & +9.4\% (.052) & +249.0\% (.394) & +111.1\% (.470) & +6.2\% (-.022) & -63.7\% (.444) & +9.5\% (.033) & -36.5\% (-.050) \\
    siamese2user & +7.6\% (.045) & +45.8\% (.116) & +45.8\% (.116) & +28.7\% (-.118) & +6.0\% (-.013) & +7.2\% (.026) & +19.8\% (.023) \\\hline\hline
    Average & +.8\% (.013) & +61.84 (.146) & +33.66\% (.134) & +8.28\% (-.035) & -6.16\% (.194) & +8.0\% (.027) & +4.7 (.001) \\
    Median & +9.4\% (.052) & +45.8\% (.116) & +4.8\% (.041) & +3.4\% (-.018) & +1.2\% (-.010) & +7.2\% (.026) & +18.6\% (.021)
    \end{tabular}
}
\caption{Performance gains of user embedding models combining with medical concepts (+Concept) comparing to standard non-concept information. We report percentage increases with absolute value increases in brackets after incorporating medical concepts. For patient relatedness, because lower is better, a negative value within brackets means a performance gain. } % We present average and median values of the five approaches that explicitly consider medical concepts.
\label{tab:plusconcept}
\end{table*}

\subsection{Intrinsic 4: patient retrieval}

% \begin{table*}[ht]
% \centering
% \begin{tabular}{c||ccccc||ccc}
% \multirow{2}{*}{Dataset} & \multicolumn{5}{c||}{Baselines} & \multicolumn{3}{c}{Our Approach} \\
%  & word2user & doc2user & suisil2user & dp2user & usr2vec & siamese2user & CAUE-GRU & CAUE-BERT \\\hline\hline
% Diabetes & 0.374 & 0.344 & 0.313 & 0.318 & 0.349 & 0.360 & \textbf{0.386} & 0.376  \\
% MIMIC-III & 0.065 & 0.132 & 0.132 & 0.113 & 0.137 & 0.116 & \textbf{0.139} & 0.132
% \end{tabular}
% \caption{Evaluation results of patient retrieval by jaccard coefficient. We highlight the best performance for each dataset. A higher score means a better performance.}
% \label{tab:retrieval}
% \end{table*}

% \begin{table}[ht]
% \centering
% \begin{tabular}{c|cc}
%     Models & Diabetes & MIMIC-III \\\hline
%     \multicolumn{3}{c}{Baselines} \\\hline
%     word2user & 0.374 & 0.065 \\
%     doc2user & 0.344 & 0.132 \\
%     suisil2user & 0.313 & 0.132 \\
%     dp2user & 0.318 & 0.113 \\
%     usr2vec & 0.349 & 0.137 \\\hline
%     \multicolumn{3}{c}{Our Approach} \\\hline
%     siamese2user & 0.360 & 0.116 \\
%     CAUE-GRU & \textbf{0.386} & 0.139 \\
%     CAUE-BERT & 0.376 & \textbf{0.142}
% \end{tabular}
% \caption{Evaluation results of patient retrieval by jaccard coefficient. We highlight the best performance for each dataset. A higher score means a better performance.}
% \label{tab:retrieval}
% \end{table}

The retrieval task is to treat each patient as a ``query'' and measure relations between the query and its retrieved top ranks.
We find the top 10 most similar patients for a query patient by calculating cosine similarities between the query patient vector and other patient vectors. 
To measure the effectiveness of user embeddings, we need to examine the quality of top-ranking returns.
Intuitively, the top returns should share similar clinical conditions with the query patient.
However, clinical labels of each patient can be more than the binary labels in previous work~\citep{amir2017quantifying} or in-hospital mortality.
The query aims to find the best candidates so that similarities between the query and the other candidates should be close to 1.
Therefore, we adapt \textit{jaccard coefficient} to measure similarity between a query patient ($u_1$) and a retrieved patient ($u_2$) by their clinical labels:

\begin{equation}
    Jaccard(u_1, u_2) = \frac{|l_1 \cap l_2|}{|l_1 \cup l_2|}
\end{equation}

We choose ten top ranks and evaluate retrieved candidates by the average Jaccard coefficients. 
We present evaluation results of the retrieval task in Table~\ref{tab:performance}.
The CAUE-GRU outperforms baselines by a large margin.
The result highlights that modeling language sequential information can be beneficial to learn patient clinical conditions for retrieval tasks.
The CAUE outperforms the siamese2user, which does not incorporate concepts.
Performance improvements show that incorporating medical concepts and leveraging document-level relations can significantly boost model performance.

\subsection{Effectiveness of Medical Concepts}

% \begin{table*}[htp]
% \centering
% \begin{tabular}{c|cc|c|cc|cc}
% \multirow{2}{*}{Models} & \multicolumn{2}{c|}{Penotype Prediction} & Mortality & \multicolumn{2}{c|}{Patient Relatedness} & \multicolumn{2}{c}{Retrieval} \\
%  & Diabetes & MIMIC-III & MIMIC-III & Diabetes & MIMIC-III & Diabetes & MIMIC-III \\ \hline\hline
% \multicolumn{8}{c}{Baselines} \\\hline
% word2user &  &  &  &  &  &  &  \\
% doc2user &  &  &  &  &  &  &  \\
% dp2user &  &  &  &  &  &  &  \\
% usr2vec &  &  &  &  &  &  &  \\ \hline\hline
% \multicolumn{8}{c}{Ours} \\\hline
% siamese2user &  &  &  &  &  &  &  \\
% CAUE-GRU &  &  &  &  &  &  &  \\
% CAUE-BERT &  &  &  &  &  &  & 
% \end{tabular}
% \caption{}
% \label{test}
% \end{table*}

We now investigate the effectiveness of medical concepts in more detail. While our approach uses neural concept embeddings, other baselines do not. 
Our main goal is to examine if medical concepts can generally improve the performance of user representations. 

We update five baselines (word2user, doc2user, dp2user, usr2vec, and siamese2user) to incorporate concept information into training user embedding under the multitask learning framework. 
For the \textit{word2user}, the model aggregates all tokens of concepts, average over the concept vectors, and combine both token and concept vectors to obtain new patient representations. For the \textit{doc2user}, we treat groups of medical concepts as documents, build a paragraph2vec~\citep{le2014distributed} model to convert concepts into vectors, and concatenate both tokens with the new concept vectors.
For the \textit{dp2user}, we build a second topic model on medical concepts besides the first topic model over tokens. We then derive concept vectors by the second topic model and train an autoencoder to reduce dimensions of the concept-topic features.
Finally, we combine the compressed concept and token vectors from the two separated autoencoders.
For the \textit{siamese2user}, we build a second prediction task to compare distances between user and concept embeddings.
For the \textit{usr2vec}, we create two joint predictions, user-token and user-concept predictions. 
Both prediction tasks follow the negative sampling process to learn user embeddings. Next, we initialize concept embeddings by averaging vectors of tokens within each concept, which gets token vectors from the pre-trained word embedding~\citep{zhang2019biowordvec}.

Table~\ref{tab:plusconcept} shows the percentage improvement and absolute value increases in brackets across four evaluation tasks when incorporating medical concepts compared to baselines without explicitly leveraging medical concepts. 
Overall, medical concepts generally augment the performance of user embeddings with an average absolute improvement of 5.93\% on Diabetes and 23.51\% on MIMIC-III respectively.
The medical concepts appear to be particularly important for CAUE, improving performance on all four evaluation tasks with an average increase in performance up to 20.3\%. 
Our close work usr2vec gains a significant improvement with an average of 40.7\%.
Comparing the different methods for incorporating medical concepts, on the one hand, our proposed CAUE-GRU works the best on average overall; on the other hand, joint learning of concepts and clinical notes can generally improve effectiveness of the baselines.

\section{Related Work}

In this work, we focus on learning user embeddings from \textbf{clinical notes}, while some studies~\citep{lei2018effective, wang2019inpatient2vec, yin2019learning, steinberg2021language} learn patient representations from tabular data.
\citeauthor{yin2019learning} applies tensor factorization to learn patient representations from tabular data, including diagnosis codes and laboratory tests.
However, learning user embeddings from clinical notes is our focus in this study, and clinical notes are noisy and unstructured data.
Several studies have proposed supervised user embeddings for clinical notes~\citep{wu2020deep, si2019deep}. 
\citeauthor{dligach2018learning} learn patient vectors by feed-forward neural networks on concept embeddings. The model averages concept vectors for each patient and predicts ICD-9 codes~\citep{dligach2018learning}.
\citeauthor{si2020patient} presents a hierarchical recurrent neural network to train embeddings for patients.
The method trains a model on MIMIC-III with phenotype annotations and applies the pre-trained model on other tasks~\citep{si2020patient}.
However, unsupervised user embeddings do not have medical annotations to learn hidden structures of patients.
Autoencoder is an encoder-decode neural network that takes patient features as inputs, compresses the features into fixed-length vectors, and reconstructs the vectors back to its input features~\citep{miotto2016deep, sushil2018patient}.
However, the autoencoder-based approaches require well-designed feature sets, and the dimension ratio between input features and user embeddings may reduce training effectiveness.
Our study fills the gap by proposing an unsupervised user embedding under a multitask framework that enforces models to recognize patients of clinical notes.
The approach does not require any human supervision and can learn user embeddings in an end-to-end style.

\textbf{Concept embeddings} aim to learn vector representations for medical concepts.
Research treats medical concepts as tokens or phrases and encodes them into vectors by contextualized word embeddings~\citep{zhang2020learning}, medical ontology~\citep{song2019medical} and graphical neural networks~\citep{choi2016multilayer, choi2020learning}.
However, few studies have incorporate medical concepts into learning patient representations under unsupervised settings.
Our study jointly leverages clinical notes and medical concepts to learn user embeddings for patients.
While our study does not focus on concept embeddings, we utilize the concept embeddings to enrich patient representations.
Recent studies utilized concepts in multimodal machine learning to align embedding spaces between vision and language~\citep{zhang2021visually}.
Our study utilizes concepts as additional constraints that balance semantic relations between patient, clinical text descriptions, and medical concepts.

\textbf{Multitask learning} simultaneously trains a model with multiple tasks.
For example, \citeauthor{si2019deep} optimize patient representations by two supervised tasks, mortality and length of stay predictions~\citep{si2019deep}.
However, this study handles a triangle relation across patient, clinical notes, and medical concepts. 
We define the relation predictions as tasks to joint optimize patient representations instead of the downstream tasks.
Our approach trains user embeddings from multiple sources and uses medical concepts to enrich patient representations.
While studies~\citep{miotto2016deep, sushil2018patient} utilize features from different sources, they concatenate patient features and train models in a single task.

\section{Conclusion}
\label{sec:end}

In this study, we have proposed a concept-aware user embedding (CAUE) in an unsupervised setting.
The model jointly leverages information of medical concepts and clinical notes using multitask learning.
The CAUE is an end-to-end neural model without any human supervision via contrastive learning and negative sampling.
Our extrinsic and intrinsic evaluations demonstrate the effectiveness of our approach, which outperforms baselines by a large margin.
Nonetheless, the medical concept shows its effectiveness in improving unsupervised baselines in general. 
We include additional implementation details in the appendix.

\paragraph{Limitation.} While experiments have demonstrated the effectiveness of our proposed unsupervised approach in clinical settings, several limitations must be acknowledged to interpret our findings appropriately.
\textit{First}, user behaviors can shift over time~\citep{han2020modelling}. 
In this study, we have proposed methods to isolate data entries by a patient per hospital visit.
The strategy can effectively reduce the temporal impacts on modeling and evaluation.
\textit{Second}, data sources may hurdle model evaluations due to their different types of clinical annotations. 
We keep evaluations consistent across the two datasets and conduct intrinsic evaluations to avoid training a large number of classifier parameters.

% temporality of user embeddings and medical events
% our features come from language only
\section*{Institutional Review Board (IRB)}
This study has no human-subject research and only uses publicly available and de-identified data, which does not need an IRB approval.

\acks{The authors thank the anonymous reviews for their insightful comments and suggestions. This work was supported in part by a research gift from Adobe Research.
The authors would also thank the HPC cluster provided by the University of Memphis.}

\bibliography{custom}

% \newpage
\appendix

\begin{table*}[ht!]
\centering
\resizebox{\textwidth}{!}{
    \begin{tabular}{cc||cc||c||cc||cc}
    \multicolumn{2}{c||}{\multirow{2}{*}{CAUE}} & \multicolumn{2}{c||}{Phenotype Prediction} & Mortality & \multicolumn{2}{c||}{Patient Relatedness} & \multicolumn{2}{c}{Retrieval} \\
    \multicolumn{2}{c||}{} & Diabetes & MIMIC-III & MIMIC-III & Diabetes & MIMIC-III & Diabetes & MIMIC-III \\\hline\hline
    \multirow{4}{*}{GRU} & None & 0.573 & 0.191 & 0.658 & 0.364 & 0.231 & 0.359 & 0.119 \\
     & +CT & 0.587 & 0.284 & 0.691 & 0.334 & 0.226 & 0.363 & 0.106 \\
     & +CC & 0.618 & 0.348 & 0.814 & 0.319 & 0.207 & 0.377 & 0.131 \\
     & +CC+CT & 0.635 & 0.369 & 0.817 & 0.292 & 0.205 & 0.386 & 0.139 \\\hline\hline
    \multirow{4}{*}{BERT} & None & 0.545 & 0.197 & 0.648 & 0.517 & 0.262 & 0.322 & 0.101 \\
     & +CT & 0.569 & 0.198 & 0.660 & 0.503 & 0.244 & 0.351 & 0.103 \\
     & +CC & 0.586 & 0.242 & 0.692 & 0.419 & 0.250 & 0.382 & 0.126 \\
     & +CC+CT & 0.593 & 0.339 & 0.792 & 0.337 & 0.235 & 0.396 & 0.142
    \end{tabular}
}
\caption{Ablation study performance. None indicates no contrastive learning (CT) and medical concept (CC). The $+$ means adding the training component. While lower scores for the patient relatedness are better, higher scores for the other tasks are better.}
\label{tab:ablation}
\end{table*}

\section{Ablation Analysis}

We compare performance impacts of individual modules in our proposed approach.
The comparison experiments follow the previous experimental settings in Section~\ref{sec:exp} and trained models with different learning rates within $[1e-4, 1e-6]$.
We experiment with the two base neural encoders, GRU and BERT, in our CAUE framework with both the contrastive learning (CT) and medical concepts (CC).
There are four combination types on top of the base neural feature extractors (GRU and BERT): 1) none of the components (None), 2) only with CT (+CT), 3) only with CC (+CC), 4) and with both optimization components (+CC+CT).
We evaluate the four combinations on the four tasks, phenotype prediction, mortality, patient relatedness, and retrieval.

We present results in Table~\ref{tab:ablation}.
Generally, learning user embeddings with CT and CC can achieve the best performance across the evaluation tasks.
Both CT and CC can improve the embedding performance.
This indicates that the two proposed strategies work coherently to improve model effectiveness on extracting patient information from clinical notes and medical concepts.
And we can also observe that medical concepts generally have more improvements over the contrastive learning.
For example, +CC models usually outperforms +CT models. 
This indicates that medical concepts may have more indicative information and correlations with the clinical codes.
% Finally, we can also find that there are higher performance variations on the BERT-based models than the GRU-based models.
% For example, in the mortality prediction task, the full BERT-based model can outperform the +CC models by near 10\%.
% We infer this as the cause of optimization uncertainties and randomness of sampling processes. 
% Finally, we can also find that GRU-based models consistently outperform than the BERT model.
% We infer this as the way to encode medical concepts: 

\section{Implementation Details}

We follow previous work~\citep{mullenbach2018explainable, harutyunyan2019multitask, stubbs2019cohort} to process MIMIC-III data.
\citeauthor{harutyunyan2019multitask} provides methods to calculate patient age information, and we wrote regular expressions to extract age information from the Diabetes corpus.
We keep clinical records of adults ($\geq 18$) in both MIMIC-III and Diabetes datasets.
The Diabetes data aggregates discharge summaries of each patient into one file, and we split individual discharge summaries by its record separator. 
To keep consistent, we use discharge summaries of both data.
We implement our models and re-implement baselines using PyTorch~\citep{adam2019pytorch}, Huggingface Transformers~\citep{wolf2020transformers}, scikit-learn~\citep{pedregosa2011scikit} and NLTK~\citep{bird2004nltk}.
For the CAUE-BERT, we experiment with multiple pre-trained BERT models~\citep{peng2019transfer, alsentzer2019publicly, lee2019biobert}. 
However, we did not find significant improvements with the different choices.
Initially, we trained models of Diabetes and MIMIC-III datasets on a NVIDIA GeForce RTX 3090, and conducted evaluation experiments on CPUs.
For the baselines, we train user2vec on the GPU and other baselines on CPUs.
We also experimented different learning rates within $[1e-4, 1e-6]$ and trained models on the university HPC cluster with NVIDIA Tesla V100s.
We then conducted the ablation analysis and reported results of the experiments.

\section{Exploratory Analysis of Medical Concepts}

Phenotype inference and patient retrieval are essential tasks for diagnosis and knowledge discovery.
Learning patient representations is critical for building robust models.
% Methods to derive patient vectors usually focus on individual tokens while not explicitly considering medical concepts~\citep{si2020patient}.
Medical concepts composed of one or more tokens in various combinations can indicate diverse semantic meanings than individual tokens.
For example, separations of ``bid'' and ``protein'' can have different semantic meanings than a combination of them, which is one type of proteins during cell death receptormediated apoptosis.
Medical concepts can correlate closely with medical annotations.
For example, ``drinker'' indicates people who drink alcohol, while ``heavy drinker'' identified in Diabetes data frequently co-occurs with the cohort annotation, ``alcohol-abuse''.
This section aims to test how the extracted concepts in our particular datasets can contribute to user modeling and how strong the effects are.

\begin{table}[htp]
\centering
\begin{tabular}{c||cc}
Dataset & Feature & Coefficient \\\hline\hline
\multirow{2}{*}{Diabetes} & n-gram & -0.037 \\
 & concept & 0.388* \\\hline
\multirow{2}{*}{MIMIC-III} & n-gram & 0.405* \\
 & concept & 0.146*
\end{tabular}
\caption{Multivariate linear regression analysis between feature and label similarities. The variables are two features n-gram and concept. * indicates that p-vale $<$ 0.05.}
\label{tab:qual}
\end{table}

To answer the question, we conduct a patient retrieval task by treating each patient as a query and retrieve similar patients. 
We define the similarity between every two patients by calculating a cosine score between their phenotype labels.
We assume that if two patients are similar, their phenotype annotations should be similar.
% Because physicians label clinical notes by reading through document descriptions, and features derived from clinical notes should be proportional to human annotations.
Therefore, if document features can accurately describe patients, the feature similarity of two patients should be proportional to their annotation similarity.

We examine the effects of concept features by a multivariate linear regression to examine relations between patient annotation and the two features types.
To represent patient features, we derive and combine two feature types, n-gram and concept.
We extract uni-, bi-, and tri-gram features and concept features that are normalized by TF-IDF.
% For each patient, we treat the patient as a query to rank the top similar patients.
We calculate feature and medical annotation (ICD-9 codes) similarities between every two patients.
Then, we can compare the two feature types by the linear regression between feature and medical annotation similarities.

We show the regression analysis results in Table~\ref{tab:qual}.
In both datasets (Diabetes and MIMIC-III), concept features show a significant correlation with medical annotations, while the n-gram features fail the significance test on the Diabetes data.
The failed test verifies the null hypothesis that there is no linear correlation between n-gram feature and medical annotation.
The results also indicate that incorporating medical concepts could help retrieve similar patients.
Concept features show a positive correlation with the medical annotation similarity.
The positive correlation means that as the annotation similarity of two patients increases, the concept feature similarity also intends to increase.
Those observations suggest there is a strong linear connection between medical concept and phenotype.

\end{document}